\documentclass{article}
\usepackage[final,eandd]{neurips_2026}
\usepackage[T1]{fontenc}
\usepackage[utf8]{inputenc}
\usepackage{microtype}
\usepackage{amsmath,amssymb}
\usepackage{graphicx,booktabs,multirow,array,tabularx}
\usepackage{enumitem,float,flafter,xcolor}
\usepackage{url}
\usepackage[hidelinks]{hyperref}
\hypersetup{pdfauthor={Han Zhang, Yilin Zhao, Zaid Pervaiz Bhat, Zheng Tang, Varun Praveen, Vidya N. Murali, David C. Anastasiu, Tomasz Kornuta},pdftitle={From Detection to Understanding: TAR and TAR-Bench for Multi-Task Traffic Anomaly Reasoning},pdfkeywords={video-language models, traffic anomaly reasoning, multi-task learning, datasets and benchmarks}}
\title{From Detection to Understanding: TAR and TAR-Bench for Multi-Task Traffic Anomaly Reasoning}
\author{%
Han Zhang$^{1}$ \quad Yilin Zhao$^{1}$ \quad Zaid Pervaiz Bhat$^{1}$ \quad Zheng Tang$^{1}$ \\
\textbf{Varun Praveen}$^{1}$ \quad
\textbf{Vidya N. Murali}$^{1}$ \quad
\textbf{David C. Anastasiu}$^{2}$ \quad
\textbf{Tomasz Kornuta}$^{1,*}$ \\
$^{1}$NVIDIA \qquad $^{2}$Santa Clara University \\
$^{*}$Corresponding Author: \texttt{tkornuta@nvidia.com}
}
\begin{document}
\maketitle
\begin{abstract}
Detecting a traffic anomaly does not establish whether a video-language model can explain what happened, localize it in time, or identify its causes. We introduce \textbf{TAR} (Traffic Anomaly Reasoning) and \textbf{TAR-Bench}, paired resources for training and evaluating these complementary capabilities across 10 tasks spanning question answering, temporal reasoning, and scene understanding. TAR contains 44,040 automatically generated annotations with chain-of-thought traces for 3,670 CCTV videos from eight public datasets. TAR-Bench provides 960 human-curated annotations for 80 held-out clips from 17 public YouTube videos. Evaluation of eleven vision-language models and analysis of fine-tuned outputs reveal temporal, causal, and descriptive errors not captured by closed-form question-answering accuracy. Training on all 10 tasks raises mean benchmark scores from 34.3 to 55.7 for Cosmos-Reason2-8B and from 30.9 to 53.9 for Qwen3-VL-8B-Instruct. These results support joint training across complementary tasks, while highlighting persistent limitations in temporal precision and causal attribution. TAR and TAR-Bench serve as the official training and in-domain evaluation resources for AI City Challenge 2026 Track~3. The dataset is available under CC-BY-4.0 at \url{https://huggingface.co/datasets/nvidia/PhysicalAI-Traffic-Anomaly-Reasoning}.
\end{abstract}
\begin{figure}[t]
  \centering
  \includegraphics[width=\textwidth]{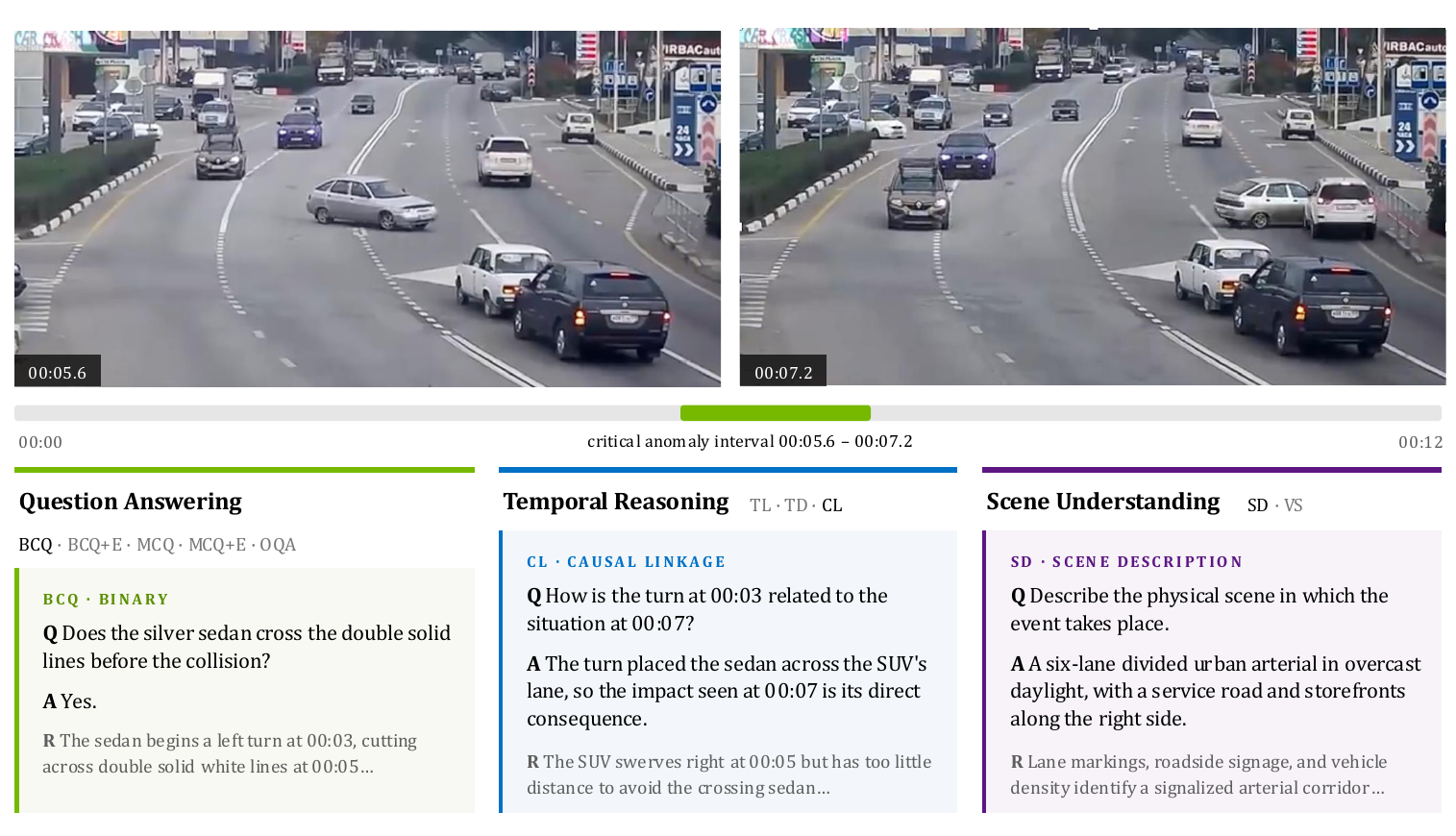}
  \caption{\textbf{TAR and TAR-Bench provide multi-task reasoning annotations for traffic videos.} Each clip is paired with questions, answers, and supporting reasoning traces for ten tasks in three groups: \emph{Question Answering}, \emph{Temporal Reasoning}, and \emph{Scene Understanding}.}
  \label{fig:example}
\end{figure}
\section{Introduction}
\label{sec:intro}

Video anomaly detection (VAD) supports traffic monitoring and road safety, but datasets such as UCF-Crime~\citep{sultani2018real}, ShanghaiTech~\citep{shanghaitech}, and TAD~\citep{tad} primarily support anomaly classification or localization. An operator also needs to understand what happened, why it happened, and how the actions of different road users contributed to the incident. Correctly answering whether a collision occurred does not establish these capabilities.

Traffic incident analysis requires these complementary capabilities. A description can identify the vehicles involved without specifying when the incident occurs. A correctly predicted interval can locate an incident without explaining the actions that led to it. Scene context is also needed to distinguish an anomalous interaction from ordinary traffic flow. A benchmark should therefore assess event recognition, temporal localization, and explanation separately, rather than treating success on one question format as comprehensive anomaly understanding.

We introduce \textbf{TAR} (Traffic Anomaly Reasoning) and \textbf{TAR-Bench}, paired training and evaluation resources with a shared 10-task taxonomy spanning \emph{Question Answering}, \emph{Temporal Reasoning}, and \emph{Scene Understanding}. TAR contains 44,040 automatically generated annotations with reasoning traces for 3,670 videos, while TAR-Bench contains 960 human-curated annotations for 80 held-out clips. This separation supports scalable training while grounding evaluation in human-corrected references.

Our contributions are:
\begin{enumerate}[leftmargin=*]
    \item \textbf{A shared multi-task formulation.} Ten tasks connect event verification to temporal, causal, and scene-level understanding through distinct answer formats and supporting reasoning traces.
    \item \textbf{Training and evaluation resources.} TAR supplies chain-of-thought supervision across eight source datasets, while TAR-Bench provides a separately human-curated test set with a documented correction process.
    \item \textbf{Empirical analysis.} Evaluation of eleven vision-language models (VLMs) exposes gaps beyond question answering. Progressive task-group ablations on two backbones show aggregate gains from multi-task fine-tuning, while diagnostic analysis identifies remaining failure modes.
    \item \textbf{Community benchmarking.} TAR and TAR-Bench serve as the official training and in-domain evaluation resources for AI City Challenge 2026 Track~3~\citep{Tang26AICity26,aicitytrack3}, supporting community evaluation of traffic anomaly reasoning (Section~\ref{sec:aicity}).
\end{enumerate}

\section{Related Work}
\label{sec:related}

We organize existing anomaly datasets and benchmarks into three broad generations: frame-level anomaly detection, event-level detection and classification, and language-grounded anomaly understanding. This progression reflects a shift from identifying anomalous events to describing and explaining them, providing the context for TAR's multi-task design.

\paragraph{Generation 1: Frame-level anomaly detection.} The earliest datasets frame anomaly detection as binary classification or frame-level localization. CUHK Avenue~\citep{cuhkavenue} provides 37 videos from a single campus camera with pixel-level anomaly masks. ShanghaiTech~\citep{shanghaitech} extends this to 13 scenes with frame-level binary labels. These datasets support identifying and localizing unusual events, but do not provide natural-language explanations of what happened or why.

\paragraph{Generation 2: Event-level detection and classification.} Larger-scale datasets introduced video-level category labels. UCF-Crime~\citep{sultani2018real} established the weakly-supervised surveillance paradigm with 1,900 videos across 13 anomaly categories (including road accidents). CADP~\citep{cadp} focuses specifically on traffic accidents with spatio-temporal annotations and time-to-accident prediction. TAD~\citep{tad} and SO-TAD~\citep{sotad} provide traffic-specific anomaly videos with binary or categorical labels. TADBenchmark~\citep{tadbenchmark} compiles traffic anomaly clips for detection benchmarking. Related work on anomaly anticipation~\citep{anastasiu2025vaa} studies predicting anomalies before they occur. These resources extend event recognition, localization, and anticipation, but do not require structured natural-language explanations across complementary reasoning tasks.

\paragraph{Generation 3: Language-grounded anomaly understanding.} This generation extends anomaly analysis to descriptions, question answering, and explicit reasoning. UCA~\citep{uca} adds 23,542 human-written, timestamped descriptions to UCF-Crime. AccidentBench~\citep{accidentbench} evaluates accident understanding through multiple-choice questions spanning temporal, spatial, and intent reasoning. SurveillanceVQA-589K~\citep{surveillancevqa} combines human captions and model-assisted QA generation across 12 question categories, including temporal, causal, and spatial understanding. Other work in this generation emphasizes explanations and structured reasoning. Holmes-VAD~\citep{holmesvad} combines anomaly judgments with natural-language explanations and manual temporal supervision, while Holmes-VAU~\citep{holmesvau} provides hierarchical clip-, event-, and video-level descriptions and analyses. VAD-R1~\citep{vadr1} provides perception-to-cognition reasoning chains and structured anomaly reports across a wider range of visual domains than TAR. TAR complements these resources with a traffic-specific formulation spanning ten tasks, combining closed-form QA, open-ended descriptions, explicit temporal intervals, and task-specific reasoning traces. It pairs automatically generated training annotations with an expert-corrected benchmark under the same task taxonomy and reports correction statistics to document the human review process.

\section{TAR and TAR-Bench}
\label{sec:dataset}

\subsection{Design Motivation}
\label{sec:design}

TAR uses the same ten task types for training and evaluation. This design allows us to examine how supervision for one task affects performance on others and whether adding task groups improves overall performance.

\paragraph{Complementary task groups.} The task groups emphasize different aspects of the same video: question answering concerns the observed event, temporal reasoning concerns its timing and progression, and scene understanding concerns the broader context. Section~\ref{sec:finetuning} evaluates this design by progressively adding task groups while testing every configuration on all ten tasks. We examine both performance on tasks included in training and transfer to tasks without direct supervision.

\paragraph{Answers and reasoning traces.} Each annotation pairs an answer with a supporting reasoning trace. This allows the same task examples to be used for answer-only supervision or for supervision that also includes reasoning. Appendix~\ref{app:reasoning_sft} compares these two approaches to examine the contribution of reasoning traces separately from the choice of training tasks.

\subsection{Source Videos}
\label{sec:videos}

TAR adds multi-task questions, answers, and reasoning traces to videos from eight public datasets (Table~\ref{tab:sources}). The training split comprises 3,670 videos ($\sim$26.1 hours): 965 anomalous and 2,705 normal. Anomalous videos contain events such as vehicle collisions, near-misses, and traffic violations, while normal videos capture routine traffic flow. The fixed-camera CCTV footage spans different geographic locations, lighting conditions, and weather.

\begin{table}[t]
  \caption{\textbf{Source datasets in TAR.} Videos span diverse traffic scenarios from CCTV surveillance cameras across multiple geographies.}
  \label{tab:sources}
  \centering
  \begin{tabular}{ll|r|r|r|r}
    \toprule
    Source Dataset & Reference & Anomalous & Normal & Total & Hours \\
    \midrule
    SO-TAD              & \citet{sotad}              & 282   & 1,904 & 2,186 & 11.2 \\
    TADBenchmark        & \citet{tadbenchmark}        & 248   & 118   & 366   & 1.7 \\
    UCF-Crime           & \citet{sultani2018real}     & 199   & 135   & 334   & 6.0 \\
    Highway Traffic     & \citet{highway}             & 0     & 254   & 254   & 0.4 \\
    TAD                 & \citet{tad}                 & 81    & 110   & 191   & 2.2 \\
    VAD-R1              & \citet{vadr1}               & 75    & 56    & 131   & 1.1 \\
    Barbados Traffic    & \citet{barbados}            & 0     & 128   & 128   & 2.1 \\
    Accident-Bench      & \citet{accidentbench}       & 80    & 0     & 80    & 1.5 \\
    \midrule
    \textbf{Total}      &                            & \textbf{965} & \textbf{2,705} & \textbf{3,670} & \textbf{26.1} \\
    \bottomrule
  \end{tabular}%
\end{table}

\subsection{Task Taxonomy}
\label{sec:tasks}

Table~\ref{tab:tasks} organizes the ten tasks into three groups. Each task specifies a question and an answer format, with a supporting reasoning trace provided as additional supervision.

\paragraph{Question Answering.} Binary questions ask whether an event occurred, multiple-choice questions require selecting the correct answer, and open-ended questions ask for a description or explanation of the anomaly. The binary and multiple-choice tasks also have variants that require an explanation with the answer. These formats assess both event recognition and the ability to explain a judgment.

\paragraph{Temporal Reasoning.} Temporal localization asks when an anomaly occurs, temporal description asks what happens within an interval, and causal linkage asks how events at two timestamps are causally related. These tasks require models to connect observations across time rather than merely recognize an event category. Their distinct outputs also allow temporal overlap and language-based descriptions to be assessed separately.

\paragraph{Scene Understanding.} Scene description and video summarization require a broader account of the environment and events. They complement local questions by asking models to synthesize context into a coherent description or summary, while preserving the identities and actions of relevant traffic participants.

\begin{table}[t]
  \caption{\textbf{Shared task taxonomy for TAR and TAR-Bench.} Each annotation includes a question, answer, and explicit chain-of-thought reasoning trace. Binary QA tasks have 2 samples per video (one positive, one negative). TAR-Bench annotations are human-curated. Answers are withheld for challenge evaluation.}
  \label{tab:tasks}
  \centering
  \resizebox{\textwidth}{!}{%
  \begin{tabular}{llllrr}
    \toprule
    Group & Task Type & Code & Description & TAR & TAR-Bench \\
    \midrule
    \multirow{5}{*}{\shortstack[l]{\textit{Question}\\\textit{Answering}}}
      & Binary QA & BCQ & Yes/No questions & 7,340 & 160 \\
      & Binary QA + Explanation & BCQ+E & Yes/No with explanation & 7,340 & 160 \\
      & Multiple-Choice QA & MCQ & Select the correct answer & 3,670 & 80 \\
      & Multiple-Choice + Explanation & MCQ+E & Select answer with explanation & 3,670 & 80 \\
      & Open-Ended QA & OQA & Question with a free-form answer & 3,670 & 80 \\
    \midrule
    \multirow{3}{*}{\shortstack[l]{\textit{Temporal}\\\textit{Reasoning}}}
      & Temporal Localization & TL & When does the anomaly occur? & 3,670 & 80 \\
      & Temporal Description & TD & What happened in the interval? & 3,670 & 80 \\
      & Causal Linkage & CL & How are events causally related? & 3,670 & 80 \\
    \midrule
    \multirow{2}{*}{\shortstack[l]{\textit{Scene}\\\textit{Understanding}}}
      & Scene Description & SD & Physical description of the scene & 3,670 & 80 \\
      & Video Summarization & VS & Summary of events in the video & 3,670 & 80 \\
    \midrule
    & \textbf{Total} & & & \textbf{44,040} & \textbf{960} \\
    \bottomrule
  \end{tabular}%
  }
\end{table}

\subsection{Annotation Pipeline}
\label{sec:pipeline}

TAR's training annotations are produced with \textbf{MAVEN} (Multi-stage Agentic Video Event aNnotation)~\citep{maven}. We summarize the TAR-specific three-stage configuration here. The pipeline architecture and adaptation procedure are described in the MAVEN paper.

\paragraph{Stage 1: Multi-scale video evidence.} Gemini 3.1 Pro~\citep{gemini3} produces a global scene caption, timestamped dense event captions, and fine-grained captions over short temporal chunks. Together these views capture scene context, event progression, and brief local evidence that a single summary can miss.

\paragraph{Stage 2: Structured event synthesis.} Gemini 3.1 Pro consolidates these captions into a Multi-Scale Spatio-Temporal Event Description (MSTED). This representation describes the overall scene and how events unfold in space and time. It also identifies an Event of Focus, including its category, cause, and consequences. For 910 of the 3,670 videos, supplementary human annotations provide global descriptions, timestamped event captions, and object boxes. These annotations are incorporated at this stage as additional evidence for the MSTED.

\paragraph{Stage 3: Multi-task generation.} Gemma-4-31B~\citep{gemma4} uses the MSTED as its sole video-specific context to generate questions, answers, and reasoning traces for the ten task types in Table~\ref{tab:tasks}. Human evidence is thus incorporated upstream, rather than supplied separately during task generation.

\paragraph{Quality control of supplementary human annotations.} Approximately 120 annotators produced the supplementary evidence for 910 training videos through annotation, quality-control, and audit stages. Appendix~\ref{app:human_correction} reports the production statistics for this workflow. These checks apply to the human-provided evidence, not to the 44,040 task annotations subsequently generated by MAVEN. A controlled audit of the generated training questions, answers, and reasoning traces remains future work.

\subsection{TAR-Bench}
\label{sec:tarbench}

TAR-Bench is the human-curated evaluation component of TAR. It contains 960 annotations across the same ten task types, covering 80 short clips trimmed from 17 public YouTube transportation videos. These clips are held out from the training corpus.

Four expert reviewers with backgrounds in vision-language models, computer vision, and traffic-anomaly analysis checked the MAVEN-generated annotations against the source videos. They corrected the content of 170 questions (17.7\%) and 356 answers (37.1\%). The reviewers divided the workload rather than independently annotating every item, so we report correction rates rather than inter-annotator agreement.

Most answer corrections changed a local factual detail rather than rewriting the response, resulting in a median string similarity of 0.95 between the original and corrected answers. Small textual changes could nevertheless alter the meaning substantially. Approximately 5--10\% of items per task required corrections that changed which agent was responsible, the causal relationship, or the interpretation of the event. These findings highlight the importance of checking benchmark references against the videos. Appendix~\ref{app:human_correction} provides the per-task statistics, counting criteria, and representative corrections. The public release includes the questions, while reference answers and reasoning traces are withheld for server-based evaluation (Section~\ref{sec:format}).

\section{Experiments}
\label{sec:experiments}

We first evaluate pretrained models on TAR-Bench, then examine the effects of adding training tasks and reasoning supervision. A diagnostic analysis complements the aggregate scores with examples of recurring errors.

\subsection{Evaluation Protocol}
\label{sec:eval}

\paragraph{Data splits.} Fine-tuning uses task-specific subsets of TAR's training corpus, with all 3,670 videos and 44,040 annotations available to the ten-task configuration. Every configuration is evaluated on all 960 TAR-Bench items from 80 held-out clips.

\paragraph{Metrics.} We report \textbf{Accuracy} [Acc] for Binary QA and Multiple-Choice QA, \textbf{Scaled BERTScore F1}~\citep{bertscore} [BS-F1] for the seven open-ended tasks, and \textbf{mean temporal intersection over union} [mIoU] for Temporal Localization. BERTScore measures semantic similarity between generated and reference answers, while temporal IoU measures overlap between predicted and reference intervals. All scores are expressed as percentages, and their unweighted \textbf{Mean} across the ten tasks is the aggregate research score. Because the tasks use different metrics, their numerical scores should not be interpreted as directly comparable measures of difficulty. BERTScore is deterministic under a fixed configuration but does not establish logical or causal correctness. Reliable evaluation of causal reasoning remains an open problem (Section~\ref{sec:discussion}).

\paragraph{Inference modes and reporting.} We evaluate models using VLMEvalKit~\citep{vlmevalkit}, with open-weight models served via vLLM~\citep{vllm}. Table~\ref{tab:zeroshot} uses direct-answer mode, without an additional chain-of-thought trace. The fine-tuning comparisons also evaluate reasoning mode, in which the model generates a chain-of-thought trace before its answer. Experimental tables report mean $\pm$ standard deviation over three runs.

\subsection{Evaluation of Pretrained VLMs}
\begin{table}[t]
  \caption{\textbf{Pretrained-model evaluation on TAR-Bench in direct-answer mode.} Values are mean $\pm$ standard deviation over three runs, expressed as percentages. Cosmos3 models are listed separately because their base-model training included TAR, so they are not zero-shot baselines for this dataset. The overall average is the unweighted mean across all eleven models.}
  \label{tab:zeroshot}
  \centering
  \setlength{\tabcolsep}{3pt}
  \resizebox{\textwidth}{!}{%
  \begin{tabular}{lccccccccccc}
    \toprule
    & \multicolumn{5}{c}{\textit{Question Answering}} & \multicolumn{3}{c}{\textit{Temporal Reasoning}} & \multicolumn{2}{c}{\textit{Scene Underst.}} & \\
    \cmidrule(lr){2-6} \cmidrule(lr){7-9} \cmidrule(lr){10-11}
    Model & \textbf{BCQ} & \textbf{BCQ+E} & \textbf{MCQ} & \textbf{MCQ+E} & \textbf{OQA} & \textbf{TL} & \textbf{TD} & \textbf{CL} & \textbf{SD} & \textbf{VS} & \textbf{Mean} \\
      & [Acc] & [BS-F1] & [Acc] & [BS-F1] & [BS-F1] & [mIoU] & [BS-F1] & [BS-F1] & [BS-F1] & [BS-F1] & [\%] \\
    \midrule
    \multicolumn{12}{l}{\textit{Proprietary}} \\
    Gemini 3.1 Pro & $64.4_{\pm 0.6}$ & $44.1_{\pm 0.6}$ & $80.4_{\pm 1.9}$ & $36.2_{\pm 1.1}$ & $30.9_{\pm 0.3}$ & $\mathbf{32.2}_{\pm 2.5}$ & $10.6_{\pm 0.4}$ & $23.5_{\pm 1.3}$ & $24.0_{\pm 1.1}$ & $16.5_{\pm 0.6}$ & $36.3_{\pm 0.3}$ \\
    \midrule
    \multicolumn{12}{l}{\textit{Open-weight (TAR-unexposed)}} \\
    VAD-R1 & $51.3_{\pm 0.0}$ & $37.8_{\pm 0.1}$ & $56.3_{\pm 0.0}$ & $32.1_{\pm 0.1}$ & $11.4_{\pm 0.1}$ & $15.1_{\pm 0.7}$ & $6.4_{\pm 0.4}$ & $18.6_{\pm 0.1}$ & $20.3_{\pm 0.0}$ & $5.3_{\pm 0.6}$ & $25.4_{\pm 0.1}$ \\
    Gemma-4-31B-IT & $56.5_{\pm 0.4}$ & $48.0_{\pm 0.3}$ & $77.5_{\pm 0.0}$ & $36.7_{\pm 0.1}$ & $27.0_{\pm 0.5}$ & $22.3_{\pm 1.5}$ & $12.4_{\pm 0.2}$ & $27.9_{\pm 0.0}$ & $15.6_{\pm 0.1}$ & $9.2_{\pm 0.5}$ & $33.3_{\pm 0.2}$ \\
    Qwen3-VL-8B & $59.6_{\pm 0.7}$ & $39.0_{\pm 0.0}$ & $62.9_{\pm 0.7}$ & $32.0_{\pm 0.3}$ & $28.8_{\pm 0.7}$ & $25.1_{\pm 0.3}$ & $11.2_{\pm 0.5}$ & $24.3_{\pm 0.3}$ & $21.9_{\pm 0.5}$ & $4.3_{\pm 0.7}$ & $30.9_{\pm 0.2}$ \\
    Qwen3-VL-32B & $65.0_{\pm 0.0}$ & $42.7_{\pm 0.2}$ & $71.2_{\pm 0.0}$ & $27.8_{\pm 0.2}$ & $13.0_{\pm 0.6}$ & $21.1_{\pm 0.3}$ & $6.6_{\pm 0.1}$ & $17.4_{\pm 0.2}$ & $14.1_{\pm 0.6}$ & $0.9_{\pm 0.2}$ & $28.0_{\pm 0.1}$ \\
    Qwen3.5-9B & $67.3_{\pm 0.4}$ & $50.7_{\pm 0.1}$ & $82.5_{\pm 0.0}$ & $70.2_{\pm 0.2}$ & $34.5_{\pm 0.1}$ & $20.4_{\pm 1.0}$ & $13.1_{\pm 0.2}$ & $17.0_{\pm 0.5}$ & $22.1_{\pm 0.1}$ & $12.8_{\pm 0.1}$ & $39.1_{\pm 0.1}$ \\
    Qwen3.5-27B & $72.5_{\pm 0.6}$ & $45.3_{\pm 0.1}$ & $85.0_{\pm 0.0}$ & $55.9_{\pm 0.4}$ & $32.7_{\pm 0.3}$ & $26.7_{\pm 0.8}$ & $14.9_{\pm 0.1}$ & $26.6_{\pm 0.2}$ & $22.0_{\pm 0.2}$ & $13.4_{\pm 0.2}$ & $39.5_{\pm 0.0}$ \\
    Cosmos-Reason2-8B & $47.7_{\pm 0.4}$ & $42.4_{\pm 0.1}$ & $65.8_{\pm 0.7}$ & $\mathbf{76.2}_{\pm 0.0}$ & $32.5_{\pm 0.3}$ & $19.8_{\pm 0.3}$ & $19.7_{\pm 0.1}$ & $12.6_{\pm 0.2}$ & $18.0_{\pm 0.2}$ & $8.3_{\pm 0.3}$ & $34.3_{\pm 0.0}$ \\
    Cosmos-Reason2-32B & $67.7_{\pm 0.4}$ & $44.5_{\pm 0.2}$ & $70.8_{\pm 0.7}$ & $53.0_{\pm 0.8}$ & $32.5_{\pm 0.0}$ & $25.1_{\pm 0.8}$ & $20.7_{\pm 0.1}$ & $26.5_{\pm 0.4}$ & $16.4_{\pm 0.2}$ & $10.2_{\pm 0.3}$ & $36.7_{\pm 0.1}$ \\
    \midrule
    \multicolumn{12}{l}{\textit{Open-weight (TAR-exposed during base-model training)}} \\
    Cosmos3-Nano & $75.8_{\pm 0.4}$ & $59.3_{\pm 0.1}$ & $86.2_{\pm 0.0}$ & $52.4_{\pm 0.2}$ & $38.7_{\pm 0.3}$ & $26.5_{\pm 0.5}$ & $25.6_{\pm 0.1}$ & $30.5_{\pm 0.1}$ & $24.9_{\pm 0.2}$ & $21.0_{\pm 0.4}$ & $44.1_{\pm 0.0}$ \\
    Cosmos3-Super & $\mathbf{89.8}_{\pm 0.7}$ & $\mathbf{60.0}_{\pm 0.2}$ & $\mathbf{88.8}_{\pm 0.0}$ & $71.2_{\pm 0.7}$ & $\mathbf{40.4}_{\pm 0.3}$ & $25.6_{\pm 0.8}$ & $\mathbf{30.3}_{\pm 0.1}$ & $\mathbf{31.0}_{\pm 0.1}$ & $\mathbf{28.8}_{\pm 0.1}$ & $\mathbf{22.4}_{\pm 0.6}$ & $\mathbf{48.8}_{\pm 0.1}$ \\
    \midrule
    \textit{Overall average} & 65.2 & 46.7 & 75.2 & 49.4 & 29.3 & 23.6 & 15.6 & 23.3 & 20.7 & 11.3 & 36.0 \\
    \bottomrule
  \end{tabular}%
  }
\end{table}

We evaluate eleven pretrained VLMs: ten general-purpose or physical-AI models and one anomaly specialist. The suite includes the proprietary Gemini 3.1 Pro~\citep{gemini3} and the open-weight Gemma-4-31B-IT~\citep{gemma4}, Qwen3-VL-8B/32B~\citep{qwen3vl}, and Qwen3.5-9B/27B~\citep{qwen35}. Cosmos-Reason2-8B/32B~\citep{cosmosreason2,cosmosreason2_32b} and Cosmos3-Nano/Super~\citep{cosmos3} target physical-AI and multimodal reasoning. VAD-R1~\citep{vadr1} is based on Qwen2.5-VL-7B and post-trained for video anomaly reasoning. Table~\ref{tab:zeroshot} distinguishes models evaluated without TAR training from Cosmos3 models, whose base-model training included TAR.

\paragraph{Overall performance and prior TAR exposure.} Cosmos3-Super achieves the highest mean score (48.8\%), while no model in the TAR-unexposed group exceeds 40\%. Qwen3.5-27B leads that group with a mean of 39.5\% and BCQ accuracy of 72.5\%. Cosmos-Reason2-32B, which was not trained on TAR, scores 36.7\%. The Cosmos3 results are not evidence of zero-shot generalization to TAR, nor do they isolate TAR's contribution from other differences in architecture and training.

\paragraph{Beyond closed-form question answering.} Strong closed-form accuracy alone does not establish comprehensive event understanding. Cosmos3-Super reaches 89.8\% on Binary QA and 88.8\% on MCQ, alongside 25.6\% temporal mIoU and BERTScore values of 30.3\% for Temporal Description, 31.0\% for Causal Linkage, 28.8\% for Scene Description, and 22.4\% for Video Summarization. These metrics assess different aspects of the response. The diagnostic analysis in Section~\ref{sec:diagnostics} further illustrates temporal, causal, and descriptive errors that closed-form accuracy does not capture.

\paragraph{Model size does not consistently predict performance.} Qwen3-VL-32B scores below its 8B counterpart (28.0\% vs.\ 30.9\%), whereas Qwen3.5-27B slightly exceeds Qwen3.5-9B (39.5\% vs.\ 39.1\%). These observations do not establish a general scaling trend: the evaluated models differ in training data and post-training, and the suite is not a controlled scaling experiment.

\paragraph{Transfer from an anomaly-specialist model.} VAD-R1 obtains a mean score of 25.4\%, compared with 30.9\% for the general-purpose Qwen3-VL-8B-Instruct baseline. It answers ``No'' on 94\% of balanced BCQ items and reaches 51.3\% accuracy, indicating a strong response bias. This example shows that specialization for one anomaly-reasoning format does not necessarily transfer to TAR's ten-task formulation.

\subsection{Finetuning on TAR, Evaluating on TAR-Bench}
\label{sec:finetuning}

We examine TAR's value for supervised fine-tuning (SFT) by progressively adding task groups. The main ablation uses Cosmos-Reason2-8B (CR2-8B), and a second ablation uses Qwen3-VL-8B-Instruct to test whether the pattern extends to another model. We compare three training configurations: (i)~\textbf{QA only} (5 \emph{Question Answering} tasks), (ii)~\textbf{QA + TR} (adding 3 \emph{Temporal Reasoning} tasks), and (iii)~\textbf{QA + TR + SC} (all 10 tasks, adding 2 \emph{Scene Understanding} tasks). Every configuration is evaluated on all ten TAR-Bench tasks in both direct-answer and reasoning modes. Table~\ref{tab:ablation} reports the CR2-8B results, and Appendix~\ref{app:qwen_task_ablation} reports the Qwen results.

\paragraph{Training setup.} We conduct supervised fine-tuning using the open-source Cosmos-RL framework~\citep{cosmosrl} with the settings reported here and the corresponding public model-configuration defaults for unspecified settings. Both backbones are fine-tuned on eight NVIDIA A100 GPUs with 80\,GB of memory each, taking approximately 40 hours per full training run of 10 tasks. For both models, we fine-tune all parameters for 3 epochs with a learning rate of $1 \times 10^{-5}$ and a global batch size of 512. Video frames are sampled at 2 frames per second (fps), with a maximum of 128 frames per video.

\paragraph{Training supervision.} Each CR2-8B training example is used twice: once with the answer only, with loss computed on answer tokens, and once with the reasoning trace and answer, with loss computed on both. This provides supervision for both inference modes. To examine the role of reasoning traces separately, we also compare answer-only and reasoning-SFT configurations on Qwen3-VL-8B-Instruct, as discussed below and in Appendix~\ref{app:reasoning_sft}.

\begin{table}[t]
  \caption{\textbf{CR2-8B task-group ablation on TAR-Bench.} SFT uses progressively more TAR task groups. The Reasoning column indicates whether chain-of-thought generation is enabled at inference. Values are mean $\pm$ standard deviation over three runs, expressed as percentages.}
  \label{tab:ablation}
  \centering
  \begin{minipage}{\linewidth}
  \setlength{\tabcolsep}{3pt}
  \resizebox{\linewidth}{!}{%
  \begin{tabular}{lcccccccccccc}
    \toprule
    & & \multicolumn{5}{c}{\textit{Question Answering}} & \multicolumn{3}{c}{\textit{Temporal Reasoning}} & \multicolumn{2}{c}{\textit{Scene Underst.}} & \\
    \cmidrule(lr){3-7} \cmidrule(lr){8-10} \cmidrule(lr){11-12}
    Model & Reasoning & \textbf{BCQ} & \textbf{BCQ+E} & \textbf{MCQ} & \textbf{MCQ+E} & \textbf{OQA} & \textbf{TL} & \textbf{TD} & \textbf{CL} & \textbf{SD} & \textbf{VS} & \textbf{Mean} \\
      &   & [Acc] & [BS-F1] & [Acc] & [BS-F1] & [BS-F1] & [mIoU] & [BS-F1] & [BS-F1] & [BS-F1] & [BS-F1] & [\%] \\
    \midrule
    \multicolumn{13}{l}{\textit{Zero-shot}} \\
    CR2-8B & Off & $47.7_{\pm 0.4}$ & $42.4_{\pm 0.1}$ & $65.8_{\pm 0.7}$ & $76.2_{\pm 0.0}$ & $32.5_{\pm 0.3}$ & $19.8_{\pm 0.3}$ & $19.7_{\pm 0.1}$ & $12.6_{\pm 0.2}$ & $18.0_{\pm 0.2}$ & $8.3_{\pm 0.3}$ & $34.3_{\pm 0.0}$ \\
    CR2-8B & On & $46.2_{\pm 0.6}$ & $6.2_{\pm 0.5}$ & $55.0_{\pm 2.2}$ & $8.4_{\pm 3.4}$ & $14.5_{\pm 0.3}$ & $13.1_{\pm 1.7}$ & $7.6_{\pm 0.6}$ & $13.4_{\pm 0.8}$ & $5.4_{\pm 0.9}$ & $2.4_{\pm 1.1}$ & $17.2_{\pm 0.5}$ \\
    \midrule
    \multicolumn{13}{l}{\textit{QA (5 tasks)}} \\
    CR2-8B & Off & $91.5_{\pm 0.3}$ & $\mathbf{64.3}_{\pm 0.3}$ & $\mathbf{96.4}_{\pm 0.0}$ & $74.3_{\pm 0.5}$ & $42.2_{\pm 0.2}$ & $15.7_{\pm 0.6}$ & $24.2_{\pm 0.2}$ & $31.6_{\pm 0.1}$ & $24.2_{\pm 0.0}$ & $31.2_{\pm 0.4}$ & $49.6_{\pm 0.0}$ \\
    CR2-8B & On & $92.3_{\pm 1.2}$ & $62.0_{\pm 0.3}$ & $92.9_{\pm 0.0}$ & $71.2_{\pm 0.3}$ & $41.5_{\pm 0.6}$ & $19.4_{\pm 1.8}$ & $28.2_{\pm 0.3}$ & $31.9_{\pm 0.4}$ & $23.1_{\pm 0.3}$ & $30.2_{\pm 0.5}$ & $49.3_{\pm 0.2}$ \\
    \midrule
    \multicolumn{13}{l}{\textit{QA + TR (8 tasks)}} \\
    CR2-8B & Off & $\mathbf{92.5}_{\pm 0.3}$ & $63.7_{\pm 0.1}$ & $95.2_{\pm 0.0}$ & $74.0_{\pm 0.2}$ & $43.1_{\pm 0.1}$ & $\mathbf{27.7}_{\pm 0.2}$ & $37.9_{\pm 0.2}$ & $\mathbf{44.5}_{\pm 0.2}$ & $25.5_{\pm 0.2}$ & $30.7_{\pm 0.6}$ & $53.5_{\pm 0.0}$ \\
    CR2-8B & On & $91.7_{\pm 1.0}$ & $61.5_{\pm 0.1}$ & $93.3_{\pm 0.7}$ & $74.1_{\pm 0.6}$ & $43.2_{\pm 0.4}$ & $24.3_{\pm 1.2}$ & $\mathbf{38.4}_{\pm 0.3}$ & $43.5_{\pm 0.9}$ & $23.6_{\pm 0.1}$ & $31.0_{\pm 0.2}$ & $52.5_{\pm 0.1}$ \\
    \midrule
    \multicolumn{13}{l}{\textit{QA + TR + SC (10 tasks)}} \\
    CR2-8B & Off & $91.7_{\pm 0.0}$ & $63.3_{\pm 0.4}$ & $91.7_{\pm 0.0}$ & $\mathbf{75.8}_{\pm 0.0}$ & $\mathbf{43.8}_{\pm 0.4}$ & $26.0_{\pm 0.3}$ & $38.1_{\pm 0.3}$ & $44.4_{\pm 0.2}$ & $\mathbf{39.2}_{\pm 0.1}$ & $\mathbf{43.4}_{\pm 0.2}$ & $\mathbf{55.7}_{\pm 0.1}$ \\
    CR2-8B & On & $91.5_{\pm 0.3}$ & $61.0_{\pm 0.4}$ & $93.7_{\pm 0.7}$ & $73.7_{\pm 0.5}$ & $42.7_{\pm 0.2}$ & $22.9_{\pm 0.8}$ & $37.0_{\pm 0.2}$ & $44.2_{\pm 0.2}$ & $35.6_{\pm 0.2}$ & $42.5_{\pm 0.4}$ & $54.5_{\pm 0.2}$ \\
    \bottomrule
  \end{tabular}%
  }
  \end{minipage}
\end{table}

We first examine task-group effects in direct-answer mode, then compare inference modes and training supervision.

\paragraph{Each task group improves aggregate performance.} Mean score increases monotonically as task groups are added: 34.3\% (zero-shot) $\rightarrow$ 49.6\% (QA) $\rightarrow$ 53.5\% (QA + TR) $\rightarrow$ \textbf{55.7\%} (all 10 tasks), a total gain of 21.4 points. This supports the utility of the combined supervision under the evaluated training configurations. Because adding task groups also adds training examples, the ablation does not isolate task diversity from data volume.

\paragraph{Transfer beyond the training tasks.} QA-only training improves Binary QA from 47.7\% to 91.5\%, Multiple-Choice QA from 65.8\% to 96.4\%, and Open-Ended QA from 32.5\% to 42.2\%. It also improves four of the five tasks without direct supervision: Temporal Description rises from 19.7\% to 24.2\%, Causal Linkage from 12.6\% to 31.6\%, Scene Description from 18.0\% to 24.2\%, and Video Summarization from 8.3\% to 31.2\%. Temporal Localization is the exception, falling from 19.8\% to 15.7\%. Thus, transfer from QA supervision is useful but does not extend uniformly to every task.

\paragraph{Benefits of task-specific supervision.} Binary QA changes little between QA-only training and the full ten-task configuration (91.5\% vs.\ 91.7\%). Adding temporal tasks raises Causal Linkage from 31.6\% to 44.5\% and Temporal Description from 24.2\% to 37.9\%. These scores remain similar after scene tasks are added (44.4\% and 38.1\%), while Scene Description and Video Summarization reach 39.2\% and 43.4\%, respectively. The full model improves Video Summarization by 35.1 points over the pretrained baseline. These results support combining task groups, although the best aggregate configuration is not the best on every individual task.

\paragraph{Effect of reasoning at inference.} Enabling chain-of-thought generation reduces the pretrained CR2-8B model's mean score from 34.3\% to 17.2\%. After ten-task fine-tuning, reasoning mode scores 54.5\%, compared with 55.7\% in direct-answer mode. Fine-tuning therefore narrows the gap between the two modes, but reasoning at inference does not improve the aggregate score for this model. These scores assess answer quality under the benchmark metrics, not the faithfulness of the reasoning traces.

\paragraph{Results on a second model.} Qwen3-VL-8B-Instruct shows the same aggregate trend. Its direct-answer mean rises from 30.9\% zero-shot to 48.6\% with five QA tasks, 51.1\% with eight QA+TR tasks, and 53.9\% with all ten tasks. The 23.0-point gain is close to the 21.4-point gain for CR2-8B. Adding temporal tasks improves Temporal Description by 8.3 points and Causal Linkage by 10.4. Adding scene tasks improves Scene Description by 14.9 points and Video Summarization by 12.0. The full results in Appendix~\ref{app:qwen_task_ablation} show that this aggregate benefit extends to a second model under the evaluated training configurations..

\paragraph{Effect of reasoning supervision.} On the ten-task Qwen configuration, answer-only SFT reaches 52.9\% in direct-answer mode, while reasoning-SFT reaches 53.9\% under the same inference setting. Enabling reasoning at inference for the latter model raises its mean to 54.4\%. This distinguishes the modest gain from training with reasoning traces from the additional effect of generating them at inference. Appendix~\ref{app:reasoning_sft} reports the per-task results. As with CR2-8B, these scores do not establish whether the generated reasoning faithfully represents the model's decision process.

\subsection{Diagnostic Error Analysis}
\label{sec:diagnostics}
To understand errors that aggregate scores can obscure, we inspect development-stage outputs from fine-tuned Cosmos-Reason2-8B and Qwen3-VL-8B. The following analysis describes four recurring failure modes, with subset statistics from the Qwen checkpoint. 

\paragraph{Anomaly omission.} Among responses whose references mention a crash, 9 of 49 scene descriptions (18.4\%) omit crash-related terms. Across all descriptive tasks, the corresponding count is 39 of 525 responses (7.4\%). In one inspected example, the reference describes a gray SUV turning into a black sedan and causing a collision, whereas the prediction describes a black SUV traveling steadily within its lane. A fluent scene description can therefore omit the central event.

\paragraph{Causal attribution.} In the Qwen checkpoint, accuracy is 79.3\% on 29 root-cause MCQ items and 90.2\% on the other 51 MCQ items. An open-ended example attributes a collision to a white car stalling, whereas the reference identifies an unsafe lane change by a black SUV. This accuracy discrepancy suggests the model is relatively good at recognizing an incident, but not correctly attributing its cause.

\paragraph{Spatial and viewpoint grounding.} Errors on spatially framed questions involve lane identity, direction, and camera-relative position. In the Qwen reasoning-mode diagnostic, error rates are 16.7\% on the spatial/viewpoint BCQ subset and 14.7\% on the corresponding MCQ subset, which is much higher than the other BCQ/MCQ items.

\paragraph{Temporal precision.} A zero-IoU prediction may lie close to the reference interval or far from it. Among the 27 zero-overlap cases in the Qwen checkpoint, 16 have a boundary gap of at most one second, and the median gap is 0.68 seconds. For short anomalies, even a small offset can eliminate overlap. Future work can consider reporting boundary distance alongside IoU, which could help distinguish these cases from larger localization errors.

\section{Data Access and Evaluation Use}
\label{sec:aicity}
\label{sec:format}
\paragraph{Public release.} The dataset card, annotations, and download scripts are available at \url{https://huggingface.co/datasets/nvidia/PhysicalAI-Traffic-Anomaly-Reasoning}. TAR annotations are distributed as ten task-specific JSON files under CC-BY-4.0. TAR-Bench combines all tasks in one file, with reference answers and reasoning traces withheld for server-based evaluation. Items include video identifiers, questions, and task-specific options or intervals. Source videos are not redistributed and retain their original terms. Retrieval scripts are provided to obtain and trim them from public sources.

\paragraph{Challenge evaluation.} AI City Challenge 2026 Track~3 uses TAR and TAR-Bench for training and in-domain evaluation~\citep{Tang26AICity26,aicitytrack3}. Its leaderboard averages nine tasks, excluding Temporal Localization, whereas this paper reports all ten tasks. Separate optional leaderboards cover fisheye traffic violations (FETV)~\citep{fisheye8k,fetv} and egocentric pedestrian intent (PSI VQA)~\citep{psi,psivqa}, offering complementary evaluation across camera geometries and reasoning targets~\citep{aicity9}. Each uses its own task definitions and metrics and is scored independently of TAR-Bench.

\section{Discussion and Limitations}
\label{sec:discussion}

\paragraph{Source video diversity.} TAR inherits the geographic, temporal, and camera biases of its eight source datasets. TAR-Bench is also limited in size: its 80 clips originate from 17 public YouTube videos. Clips from the same source may share scene characteristics, so they should not be treated as fully independent samples of traffic conditions. The challenge's out-of-domain evaluations provide complementary tests under different camera views and task settings.

\paragraph{Domain specificity.} TAR's released configuration and annotations cover transportation video, predominantly from fixed surveillance cameras. Although MAVEN can be adapted to other domains and question styles~\citep{maven}, that transfer is not evaluated here.

\paragraph{Coverage of rare anomalies.} No finite collection can represent every type of traffic anomaly. Before adopting the current data strategy, we explored physics simulators including CARLA~\citep{carla} and video-to-video translation to make the simulated footage more realistic. A persistent simulation-to-reality gap limited transfer to surveillance footage. Combining simulation with real-scene evidence may improve coverage of rare events, but this remains future work.

\paragraph{Annotation quality.} TAR's 44,040 training annotations are generated by VLMs and may contain hallucinations, omissions, or errors in timing, location, and attribution. Human-provided evidence for 910 videos undergoes quality control, but the generated task labels have not undergone a controlled error-rate or agreement study. TAR-Bench receives a separate expert review, as documented in Section~\ref{sec:tarbench}. That review does not validate the training labels, which should be treated as automatically generated supervision with residual noise rather than human-verified ground truth.

\paragraph{Model coverage.} The VAD-R1 experiment provides an anomaly-specialist comparison, but one specialist cannot characterize the broader anomaly-model landscape. The task-group ablations show an aggregate benefit on two models, not generalization to arbitrary architectures or video domains. They also vary training-data volume alongside task diversity.

\paragraph{Evaluating causal reasoning.} BERTScore measures semantic similarity, not logical correctness: a response can resemble a reference while assigning the wrong cause or reversing an event relation. Reliable evaluation of open-ended causal reasoning remains an open problem. Causal Linkage scores should therefore be interpreted as similarity to reference explanations, not as validated measures of causal reasoning. One possible approach is to assess responses through explicit binary questions about event occurrence, the responsible agent, actions, temporal order, and causal relations, following work on interpretable LLM evaluation~\citep{askdontjudge}. Such a rubric would require validation against human judgments for traffic video and is not used in the reported results.

\paragraph{Broader impacts.} Traffic reasoning may support road safety and incident analysis, but can also enable harmful surveillance or erroneous attribution. Public availability does not eliminate privacy risks or supersede source licenses. We release annotations rather than source videos and document source provenance and intended use. These resources do not establish readiness for autonomous safety-critical or enforcement decisions.

\section{Conclusion and Future Work}
\label{sec:conclusion}

TAR and TAR-Bench pair 44,040 automatically generated training annotations with 960 human-curated evaluation annotations across ten traffic anomaly reasoning tasks. Their shared taxonomy covers question answering, temporal reasoning, and scene understanding, extending evaluation beyond detecting an anomaly to describing its context, locating it in time, and explaining relationships between events.

Progressive multi-task fine-tuning improves aggregate scores by 21.4 points on Cosmos-Reason2-8B and 23.0 points on Qwen3-VL-8B-Instruct. Despite these gains, models still omit important events, misattribute causes, and make spatial and temporal errors. The results support complementary task supervision and highlight the need to interpret aggregate scores alongside task-specific diagnostics. Future work should broaden video sources, audit the generated training labels, and develop more reliable measures of causal and temporal correctness.

\clearpage
\begin{ack}
We thank Kha Le, Danh Nguyen, Huy Bui, Quynh Ta, and Hien Mai for their assistance with video sourcing. We thank Arihant Jain for help with the annotation specification and evaluation of human annotation quality, Suseella Panguluri and the NVIDIA Data Factory team for contextual labeling support, and Bhanu Pisupati for discussions and evaluation feedback. For the AI City Challenge 2026 Track~3 out-of-domain evaluations, we thank Munkhjargal Gochoo and Ahmed Abduljawad from United Arab Emirates University for contributing the FishEye Traffic Violation benchmark; and we thank Renran Tian and Shaozhi Wang from North Carolina State University for contributing the Egocentric Dashcam Pedestrian Intent benchmark.

\paragraph{Funding and competing interests.} The authors report no third-party funding or support for this work. Seven authors are affiliated with NVIDIA, which develops the Cosmos-Reason2 and Cosmos3 models evaluated here. The authors report no additional competing interests beyond their stated institutional affiliations.
\end{ack}

\bibliographystyle{plainnat}
\bibliography{references}
\clearpage
\appendix
\raggedbottom

\section{Human Correction Analysis}
\label{app:human_correction}

This appendix details the expert corrections summarized in Section~\ref{sec:tarbench}. We report the per-task counts, explain the types of changes, and provide representative examples. The final paragraph describes the separate quality-control process for human-provided evidence used in training-data generation.

\paragraph{Review process and correction counts.} Four experts divided the TAR-Bench annotations and checked each assigned item against its source video. Table~\ref{tab:human_corrections} counts changes to question or answer content. Changes limited to formatting, standard answer instructions, whitespace, or reordering multiple-choice options without changing their content are excluded. Question and answer corrections are counted separately and may occur in the same item. These are correction statistics, not inter-annotator agreement measurements.

\begin{table}[H]
  \caption{\textbf{Complete human-correction statistics for the 960 TAR-Bench items.} Percentages in the total row are relative to all items.}
  \label{tab:human_corrections}
  \centering
  \setlength{\tabcolsep}{7pt}
  \begin{tabular}{l|r|r|r}
    \toprule
    Task & Items & Questions corrected & Answers corrected \\
    \midrule
    BCQ   & 160 & 38 & 0 \\
    BCQ+E & 160 & 29 & 53 \\
    MCQ   & 80  & 30 & 21 \\
    MCQ+E & 80  & 31 & 19 \\
    OQA   & 80  & 5  & 46 \\
    TL    & 80  & 24 & 11 \\
    TD    & 80  & 3  & 46 \\
    CL    & 80  & 8  & 49 \\
    SD    & 80  & 0  & 53 \\
    VS    & 80  & 2  & 58 \\
    \midrule
    \textbf{Total} & \textbf{960} & \textbf{170 (17.7\%)} & \textbf{356 (37.1\%)} \\
    \bottomrule
  \end{tabular}
\end{table}

\paragraph{Extent of answer changes.} Most corrected answers retain their original structure and require a local factual correction rather than a rewrite. The median string similarity between original and corrected answers is 0.95, and mean answer length is nearly unchanged. However, roughly 5--10\% of items per task require changes that alter the responsible agent, causal direction, or interpretation of the event. This smaller group illustrates why high string similarity does not establish semantic correctness.

\paragraph{Types of corrections.} Question edits primarily address misidentified objects or attributes, incorrect timestamps, and multiple-choice distractors that require substantive rewriting. Answer edits mostly correct entity identities or fine-grained attributes within otherwise usable descriptions. A smaller set changes the causal agent, temporal relation, or justification. For example, 38 BCQ questions required entity-identity corrections while their Yes/No answers remained unchanged. Table~\ref{tab:correction_examples} illustrates both local factual corrections and changes with a larger effect on meaning.

\begin{table}[H]
  \caption{\textbf{Representative question and answer corrections.} Examples illustrate changes to object identity, attributes, temporal grounding, causal attribution, and justification.}
  \label{tab:correction_examples}
  \centering
  \small
  \setlength{\tabcolsep}{4pt}
  \begin{tabular}{p{0.2\textwidth}|p{0.36\textwidth}|p{0.36\textwidth}}
    \toprule
    Category & Original & Corrected \\
    \midrule
    Object identity & Does a silver sedan collide with the black SUV? & Does a silver hatchback collide with the black SUV? \\ \midrule
    Fine-grained attribute & The light pink car approaches the intersection. & The light beige car approaches the intersection. \\ \midrule
    Timestamp grounding & Explain the relationship between the event at 00:02 and the situation at 00:04. & Explain the relationship between the event at 00:04 and the situation at 00:07. \\ \midrule
    Semantic reversal & The root cause is the black sedan's failure to obey traffic signals. & The root cause is the white sedan's failure to obey traffic signals. \\ \midrule
    Justification replacement & No. The traffic light is green throughout the video. & No. A black sedan collides with the back of a slowing white sedan in the leftmost lane. \\
    \bottomrule
  \end{tabular}
\end{table}

\paragraph{Separate quality control for training-video evidence.} Approximately 120 annotators produced the supplementary human annotations for 910 training videos through annotation, quality-control, and audit stages. Items that failed review were returned for revision. Production statistics were 99.8\% first-pass acceptance, 100\% acceptance after revision, and 99.9\% agreement between quality control and audit. These figures describe the human-provided evidence incorporated into MAVEN's event synthesis, not the 44,040 generated task annotations. They therefore do not estimate the error rate of the automatically labeled training corpus.

\section{Qwen3-VL-8B Task-Group Ablation}
\label{app:qwen_task_ablation}

This ablation tests whether the task-group trend observed for CR2-8B also appears on Qwen3-VL-8B-Instruct. Training progresses from five Question Answering tasks (QA), to eight tasks with Temporal Reasoning (QA+TR), and then all ten tasks with Scene Understanding (QA+TR+SC). Every configuration is evaluated on all ten TAR-Bench tasks. The Reasoning column in Table~\ref{tab:qwen_ablation} refers to chain-of-thought generation at inference, not the choice of training tasks. The ten-task fine-tuned configuration is also reported as reasoning-SFT in Appendix~\ref{app:reasoning_sft}, where it is compared with answer-only supervision.

In direct-answer mode, the mean score rises from 30.9\% zero-shot to 48.6\%, 51.1\%, and 53.9\% as task groups are added. This aggregate improvement is not uniform across individual tasks: Temporal Localization peaks with eight tasks, while the full configuration yields the highest Scene Description and Video Summarization scores. Reasoning at inference is shown to increase the ten-task mean from 53.9\% to 54.4\%, but the gains vary by tasks and are more prominent on BCQ, MCQ+E, SD and VS tasks.

\begin{table}[H]
  \caption{\textbf{Qwen3-VL-8B-Instruct task-group ablation on TAR-Bench.} Reasoning indicates chain-of-thought generation at inference. Values are mean $\pm$ standard deviation over three runs, expressed as percentages.}
  \label{tab:qwen_ablation}
  \centering
  \setlength{\tabcolsep}{2.6pt}
  \resizebox{\textwidth}{!}{%
  \begin{tabular}{lcccccccccccc}
    \toprule
    Training & Reasoning & BCQ & BCQ+E & MCQ & MCQ+E & OQA & TL & TD & CL & SD & VS & Mean \\
    \midrule
    Zero-shot & Off & $59.6_{\pm 0.7}$ & $39.0_{\pm 0.0}$ & $62.9_{\pm 0.7}$ & $32.0_{\pm 0.3}$ & $28.8_{\pm 0.7}$ & $25.1_{\pm 0.3}$ & $11.2_{\pm 0.5}$ & $24.3_{\pm 0.3}$ & $21.9_{\pm 0.5}$ & $4.3_{\pm 0.7}$ & $30.9_{\pm 0.2}$ \\\midrule
    QA (5) & Off & $86.9_{\pm 0.0}$ & $60.2_{\pm 0.2}$ & $82.5_{\pm 0.0}$ & $74.7_{\pm 0.1}$ & $42.2_{\pm 0.3}$ & $29.6_{\pm 0.5}$ & $25.7_{\pm 0.2}$ & $30.1_{\pm 0.2}$ & $26.1_{\pm 0.3}$ & $28.1_{\pm 0.3}$ & $48.6_{\pm 0.1}$ \\
    QA (5) & On & $87.7_{\pm 0.4}$ & $60.5_{\pm 0.4}$ & $77.1_{\pm 0.7}$ & $72.0_{\pm 0.7}$ & $41.8_{\pm 1.1}$ & $30.0_{\pm 1.0}$ & $25.8_{\pm 0.4}$ & $28.0_{\pm 0.4}$ & $24.0_{\pm 0.2}$ & $27.6_{\pm 0.4}$ & $47.4_{\pm 0.2}$ \\ \midrule
    QA+TR (8) & Off & $89.6_{\pm 0.4}$ & $60.1_{\pm 0.1}$ & $86.3_{\pm 0.0}$ & $73.9_{\pm 0.1}$ & $41.6_{\pm 0.2}$ & $\mathbf{30.6}_{\pm 0.3}$ & $34.0_{\pm 0.2}$ & $40.5_{\pm 0.2}$ & $24.4_{\pm 0.3}$ & $30.0_{\pm 0.4}$ & $51.1_{\pm 0.0}$ \\
    QA+TR (8) & On & $90.8_{\pm 1.0}$ & $59.3_{\pm 0.2}$ & $84.6_{\pm 0.7}$ & $75.1_{\pm 0.6}$ & $\mathbf{43.2}_{\pm 0.8}$ & $29.3_{\pm 0.4}$ & $34.5_{\pm 0.3}$ & $40.7_{\pm 0.3}$ & $23.2_{\pm 0.2}$ & $29.8_{\pm 0.7}$ & $51.1_{\pm 0.1}$ \\ \midrule
    QA+TR+SC (10) & Off & $87.7_{\pm 0.4}$ & $\mathbf{60.7}_{\pm 0.2}$ & $\mathbf{87.1}_{\pm 0.7}$ & $75.4_{\pm 0.0}$ & $42.6_{\pm 0.4}$ & $28.2_{\pm 0.3}$ & $\mathbf{35.1}_{\pm 0.2}$ & $\mathbf{41.2}_{\pm 0.3}$ & $39.3_{\pm 0.3}$ & $42.0_{\pm 0.3}$ & $53.9_{\pm 0.2}$ \\
    QA+TR+SC (10)  & On & $\mathbf{91.3}_{\pm 0.6}$ & $59.8_{\pm 0.2}$ & $84.6_{\pm 1.9}$ & $\mathbf{76.1}_{\pm 0.6}$ & $43.0_{\pm 0.2}$ & $28.6_{\pm 0.4}$ & $34.9_{\pm 0.5}$ & $\mathbf{41.2}_{\pm 0.6}$ & $\mathbf{41.0}_{\pm 0.4}$ & $\mathbf{43.1}_{\pm 0.4}$ & $\mathbf{54.4}_{\pm 0.1}$ \\ 
    \bottomrule
  \end{tabular}}
\end{table}

\clearpage
\section{Answer-Only versus Reasoning-SFT}
\label{app:reasoning_sft}

This comparison keeps the ten-task Qwen3-VL-8B-Instruct configuration fixed and varies the training supervision. Answer-only SFT uses the answers without their supporting reasoning traces, while reasoning-SFT includes reasoning supervision. Table~\ref{tab:reasoning_sft} reports both models in direct-answer mode and additionally evaluates reasoning-SFT with chain-of-thought generation at inference.

Under direct-answer inference, reasoning-SFT raises the mean from 52.9\% to 53.9\%. Enabling reasoning at inference raises the latter model's mean to 54.4\%. The gains are modest and vary by task. For example, answer-only SFT achieves the highest Temporal Localization score in this comparison. These results distinguish the effect of reasoning supervision from the effect of reasoning at inference, but do not establish the faithfulness of the generated traces.

\begin{table}[H]
  \caption{\textbf{Full answer-only versus reasoning-SFT comparison on TAR-Bench.} Values are mean $\pm$ standard deviation over three runs, expressed as percentages.}
  \label{tab:reasoning_sft}
  \centering
  \setlength{\tabcolsep}{2.4pt}
  \resizebox{\textwidth}{!}{%
  \begin{tabular}{lcccccccccccc}
    \toprule
    Training configuration & Reasoning & BCQ & BCQ+E & MCQ & MCQ+E & OQA & TL & TD & CL & SD & VS & Mean \\
    \midrule
    Zero-shot & Off & $59.6_{\pm 0.7}$ & $39.0_{\pm 0.0}$ & $62.9_{\pm 0.7}$ & $32.0_{\pm 0.3}$ & $28.8_{\pm 0.7}$ & $25.1_{\pm 0.3}$ & $11.2_{\pm 0.5}$ & $24.3_{\pm 0.3}$ & $21.9_{\pm 0.5}$ & $4.3_{\pm 0.7}$ & $30.9_{\pm 0.2}$ \\ \midrule
    10-task answer-SFT & Off & $88.1_{\pm 0.0}$ & $59.5_{\pm 0.1}$ & $83.8_{\pm 0.0}$ & $72.3_{\pm 0.1}$ & $42.8_{\pm 0.2}$ & $\mathbf{31.6}_{\pm 0.3}$ & $32.2_{\pm 0.3}$ & $38.7_{\pm 0.2}$ & $38.7_{\pm 0.3}$ & $41.3_{\pm 0.1}$ & $52.9_{\pm 0.1}$ \\ \midrule
    10-task reasoning-SFT & Off & $87.7_{\pm 0.4}$ & $\mathbf{60.7}_{\pm 0.2}$ & $\mathbf{87.1}_{\pm 0.7}$ & $75.4_{\pm 0.0}$ & $42.6_{\pm 0.4}$ & $28.2_{\pm 0.3}$ & $\mathbf{35.1}_{\pm 0.2}$ & $\mathbf{41.2}_{\pm 0.3}$ & $39.3_{\pm 0.3}$ & $42.0_{\pm 0.3}$ & $53.9_{\pm 0.2}$ \\
    10-task reasoning-SFT & On & $\mathbf{91.3}_{\pm 0.6}$ & $59.8_{\pm 0.2}$ & $84.6_{\pm 1.9}$ & $\mathbf{76.1}_{\pm 0.6}$ & $\mathbf{43.0}_{\pm 0.2}$ & $28.6_{\pm 0.4}$ & $34.9_{\pm 0.5}$ & $\mathbf{41.2}_{\pm 0.6}$ & $\mathbf{41.0}_{\pm 0.4}$ & $\mathbf{43.1}_{\pm 0.4}$ & $\mathbf{54.4}_{\pm 0.1}$ \\
    \bottomrule
  \end{tabular}}
\end{table}

\end{document}